\documentclass{article}
\usepackage{ijcai23}

\usepackage{times}
\usepackage{soul}
\usepackage{url}
\usepackage[hidelinks]{hyperref}
\usepackage[utf8]{inputenc}
\usepackage[small]{caption}
\usepackage{graphicx}
\usepackage{amsmath}
\usepackage{amsthm}
\usepackage{booktabs}
\usepackage[ruled]{algorithm2e}
\usepackage{booktabs,makecell}
\usepackage{multicol}
\usepackage{multirow}

\usepackage[switch]{lineno}
\usepackage{tcolorbox}
\usepackage{soul}
\usepackage[T1]{fontenc}
\usepackage[utf8]{inputenc}
\usepackage[font=small]{caption}
\usepackage{hyperref}
\hypersetup{hidelinks}
\usepackage{subfigure}
\linenumbers

\title{ReOnto: A Neuro-Symbolic Approach for Biomedical Relation Extraction}

\author{
   Monika Jain$^1$
   \and
   Kuldeep Singh$^2$ \and
   Raghava Mutharaju$^1$
    \affiliations
$^1$Knowledgeable Computing and Reasoning (KRaCR) Lab, IIIT-Delhi, India\\
$^2$Cerence GmbH and Zerotha Research, Germany\\    \emails
\{monikaja, raghava.mutharaju\}@iiitd.ac.in\\kuldeep.singh1@cerence.com}

\begin{document}
\nolinenumbers
\maketitle

\begin{abstract}
    Relation Extraction (RE) is the task of extracting semantic relationships between entities in a sentence and aligning them to relations defined in a vocabulary, which is generally in the form of a Knowledge Graph (KG) or an ontology. Various approaches have been proposed so far to address this task. However, applying these techniques to biomedical text often yields unsatisfactory results because it is hard to infer  relations directly from sentences due to the nature of the  biomedical relations. To address these issues, we present a novel technique called ReOnto, that makes use of neuro symbolic knowledge for the RE task. ReOnto employs a graph neural network to acquire the sentence representation and leverages publicly accessible ontologies as prior knowledge  to identify the sentential relation between two entities. The approach involves extracting the relation path between the two entities from the ontology. We evaluate the effect of using symbolic knowledge from ontologies with graph neural networks. Experimental results on two public biomedical datasets,  BioRel and ADE, show that our method outperforms all the baselines (approximately by 3\%). 

\end{abstract}

\section{Introduction}

In recent times, due to the exponential increase in data, knowledge bases have gained popularity as a means to efficiently store and organize information~\cite{fensel2020we}. Although considerable efforts are invested in updating and maintaining knowledge bases, their incompleteness persists due to the dynamic nature of facts, which constantly evolve over the Web and other sources. Hence, there is a need to automate the process of extracting knowledge from text. Relation Extraction (RE) is task of predicting the relation given a sentence and an entity pair \cite{bastos2021recon}. In domains such as biomedicine, relation extraction task poses a few critical domain-specific challenges. Consider a sentence, \textit{atrio ventricular (C0018827) conduction defects and arrhythmias by selective perfusion of a-v conduction system in the canine heart (C0018787)}, with entities C0018827 (ventricular) and C0018787 (heart) linked to UMLS \cite{bodenreider2004unified}.  Here target relation is \textit{hasPhysicalPartOfAnatomicStructure}. The RE task aims to infer the semantic relationships. As demonstrated in the example, working with biomedical corpora poses several challenges. These include: complex input sentences that may require extensive parsing and interpretation to extract relevant information. Indirectly inferred relations between entities in the text, which may require sophisticated natural language processing techniques. Difficulty obtaining domain knowledge of the specific entities mentioned in the text, which may require specialized expertise and additional research. Moreover, in the biomedical domain, entities are intricately interlinked, resulting in numerous densely linked entities with high degrees and multiple paths connecting them~\cite{angell2021clustering}. Hence, inferring the correct relation from a given sentence may require  reasoning about the potential path.

\textbf{Limitation of Existing Works and Hypothesis}. The existing approaches employ various techniques for relation extraction such as multi-task learning \cite{crone2020deeper}, transformers \cite{eberts2020span}, Graph Neural Network (GNN) models \cite{bastos2021recon,zhu-etal-2019-graph} have been used to process complex relationships between entities. While deep learning models \cite{nadgeri-etal-2021-kgpool,santosh2021joint} can incorporate semantic information of entities. Albeit effective, these models employ standard message-passing or attention-based
approaches (transformers, GNNs) which are inherently focused on homophilic signals \cite{balcilar2020analyzing,bastos2022how} (i.e., only on neighborhood interactions) and ignore long-range interactions that may be required to infer the semantic relationship between two biomedical entities. Furthermore, sufficient domain-specific knowledge is available in various biomedical ontologies to be used as background knowledge for relation extraction. It is also evident in the literature that reasoning over ontologies \cite{bona_enhancing_2019,winnenburg_using_2015} allow capturing long-range dependencies between two entities \cite{pan2019entity,zhang2022knowledge}, which further helps in making predictions. For instance, in \cite{hong_ontology-based_2004}, ontology information was utilized as a tuple and transformed into a 3-D vector for predicting compound relations.
Hence, it remains an open \textbf{research question}: for biomedical relation extraction, can we combine reasoning ability over publicly available biomedical ontologies to enrich an underlying deep learning model which is inherently homophilic? 

\textbf{Contributions:} To tackle this research question, to our knowledge our approach represents the first neuro-symbolic method for extracting relations in the biomedical domain. Our method is two-fold. Firstly, we aim to aggregate the symbolic knowledge in the form of  axioms (facts) consisting of logical constructs and  quantifiers such as  \textit{there exist}, \textit{for all}, \textit{union} and  \textit{intersection} between entities present in various public ontologies and build background knowledge. In the second step, we incorporate background knowledge into a Graph Neural Network (GNN) to enhance its capabilities to capture long-range dependencies. The rationale behind using a GNN is to exploit the correlations between entities and predicates due to its message-passing ability between the nodes. Inducing external symbolic knowledge makes our approach transparent as we can backtrack the paths used for inducing long-range dependencies between entities. Hence, we empower the GNN by externally induced symbolic knowledge to capture long-range interactions needed to infer biomedical relations between two given entities and a sentence. We name our approach as ``ReOnto" containing following key contributions. 
\begin{itemize}

  \item Our novel relation extraction method, ReOnto, utilizes an ontology model to learn subgraphs containing expressive axioms connecting the given entities. It consists of a symbolic module incorporating domain-specific knowledge into a GNN, enabling the prediction of required relations between two entities within a biomedical knowledge graph.
  \item  We study the effect of symbolic knowledge on the performance of the underlying deep learning model by considering several key characteristics such as 1) entity coverage from ontology, 2)  the number of hops, etc. We provide conclusive evidence that aggregating knowledge from various sources to build the symbolic component (instead of using just one ontology for background knowledge) has a positive impact on the overall performance. 
  \item We provide an exhaustive evaluation on two standard datasets, and our proposed method outperforms all baselines for biomedical relation extraction. 
\end{itemize}

\section{Related work}
\textbf{Muti-instance RE}: Multi-instance relation extraction aims to utilize previous mentions of entities in a given document to infer the semantic relationship between them. Some approaches leverage attention-based convolution neural network~\cite{shen-huang-2016-attention}, multi-level CNN attention~\cite{wang-etal-2016-relation} and by ranking with CNN to classify relation~\cite{https://doi.org/10.48550/arxiv.1504.06580}. In contrast, alternative approaches employ recurrent neural networks for relation classification~\cite{https://doi.org/10.48550/arxiv.1508.01006} and hierarchical RNN with attention. Besides this, some works also use entity context information such as type and descriptions to improve the performance~\cite{vashishth-etal-2018-reside}. To deal with the noise at the sentence-level and bag level, \cite{https://doi.org/10.48550/arxiv.1904.00143} proposed a distant supervision approach incorporating intra-bag and inter-bag attentions.

\textbf{Sentential and Biomedical RE}:
GP-GNN~\cite{zhu-etal-2019-graph} proposed a graph neural network with generated parameters which solves the relational message-passing task by encoding natural language as parameters and performing propagation from layer to layer. RECON~\cite{bastos2021recon} is an extended approach which uses the entity details like alias, labels, description and instance in an underlying GNN model for sentential RE. As discussed in \cite{nadgeri-etal-2021-kgpool}, not all facts contribute to improved performance, and therefore, the context must be dynamically selected based on the given sentence. However, these works are limited to general domain and finds their limitation in the biomedical domain.
In the biomedical domain, \cite{crone2020deeper} introduced a multi-task learning approach that utilizes joint signals from entity extraction task to improve relation extraction. 
\cite{santosh2021joint} enriched the performance of biomedical relation extraction by incorporating linguistic information and entity types into a BERT model. \cite{cabot2021rebel} employed an end-to-end seq2seq model for biomedical RE. 

\textbf{Ontology based RE}:
The authors of reference~\cite{ontology} proposed using an ontology as a hyperlink structure for the web to facilitate relation extraction.
 Authors utilize the web structure using a breadth-first search for relation extraction.~\cite{aghaebrahimian2022ontology} uses RNN with a convolutional neural network to process three features: tokens, types, and graphs. Work claim that entity type and ontology graph structure provide better representations than simple token-based representations for RE. We point readers to~\cite{karkaletsis_ontology_2011} for details on ontology-powered information systems.


\section{Problem Formulation and Approach}
\begin{figure}[ht]
\centering
\includegraphics[width=8.5cm]{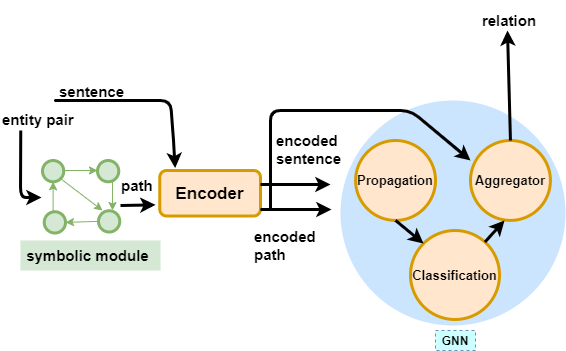}
\caption{ReOnto Approach. The role of the symbolic module is to aggregate symbolic knowledge. It takes the entity pair and gives path information. 1) Encoding module accepts input vectors of sentence and path information to provide transition matrix. 2) Propagation module shares the hidden states of generated transition matrix with its neighbors 3) Classification module provides scores of prediction 4) Aggregator module integrates the  score of the biased relation (from ontology reasoning) with that of the one  from GNN to calculate loss.}
\label{fig:archi}
\end{figure}

 We define a KG as a tuple $KG = (\mathcal{E},\mathcal{R},\mathcal{T}^+)$ where $\mathcal{E}$ denotes the set of entities (vertices), $\mathcal{R}$ is the set of relations (edges), and $\mathcal{T}^+ \subseteq \mathcal{E} \times \mathcal{R} \times \mathcal{E} $ is a set of all triples. The \textit{RE Task} aims to find the target relation $r^c \in \mathcal{R}$ for a given pair of entities $\langle e_i,e_j\rangle$ within the sentence $\mathcal{W}$. If no relation is inferred, it returns \textit{NA} label.
 In this section, we first discuss the ReOnto framework, which integrates the power of the graph neural network (GNN) \cite{bastos2021recon} with that of symbolic knowledge. A GNN primarily employs three modules, which are  encoding, propagation, and classification. Symbolic knowledge is integrated with the GNN score in the aggregation module (Figure \ref{fig:archi}).

\subsection{Symbolic Module}
As a first step, we aggregate symbolic knowledge(SK), available in public ontologies for extracting long-range dependencies between entity pairs. We build a connected graph G of the symbolic knowledge derived from ontologies. We define $\textrm{G = (V, E, T}_+$) where V has a set of entities such that each edge (v$_s$, v$_o$) $\in$ E and v$_s$, v$_o$ $\in$ V corresponds to a sequence s = s$_0^{s,o} s_1^{s,o}s_2^{s,o}$ $\ldots$ $s_{l-1}^{s,o}$  extracted from the  text where s, o represent the source and destination entities. 
We also consider related SK of entity pair SK$^{s,o}$ which consist of path information ($\sum$ $path_0^{i}$;$\sum$ $axiom\_path_0^{i}$) $\in SK^{s,o}$ where i is the number of hops traversed to get the path. Path consists of multi hop details, each containing detailed information, while the axiom path contains path information enriched with expressive axioms.
We identify the, directly and indirectly, connecting paths between the entities v$_s$ and v$_o$ (Algorithm \ref{algo}).
\begin{algorithm}[ht]
  \caption{Path generation via ontology}
  \label{algo}
  
  \SetKwInOut{KwIn}{Input}
  \SetKwInOut{KwOut}{Output}
  \SetKwProg{Fn}{Function}{:}{}
  \SetKwFunction{path}{PathGeneration}
  \SetKwFunction{FMain}{ExplorePath}
  \SetKwFunction{FMein}{ExploreSymbolicPath}
  
  \KwIn{entity pair (v$_s$,v$_o$), Number of hops (N)}
  \KwOut{finalpath}
  \DontPrintSemicolon
  
  \textbf{Initialization:} \\
  \hspace{0.5cm} $i=1$, source= v$_s$, path$_{i-1}$, axiom\_path$_{i-1}$, finalpath, adjacent node, hop\_path$_i$, axiom\_path$_i$ =  \{\} \\
finalpath = PathGeneration(v$_s$, v$_o$, N)
  
  \Fn{\path{v$_s$, v$_o$, N}}
  {
    path, axiom\_path, finalpath =  \{\} \\
    
    \ForEach{entity pair $v_s, v_o \in$ ontology}{
      path.append(ExplorePath($v_s, v_o$, N))\;
      axiom\_path.append(ExploreSymbolicPath($v_s, v_o$, N))\;
    }

    finalpath = path $\cup$ axiom\_path\;
    \KwRet \{finalpath\} \;
  }
  
  \Fn{\FMain{v$_s$, v$_o$, $N$}}
  {
    hop\_path$_i$, adjacent node= GetNHopFromSource(v$_s$, 1) {\hfill\textbf{//}\,calculates 1 hop distance from source}\;
    path$_{i}$ = hop\_path$_i$ $\cup$ path$_{i-1}$ \;
    \If{$v_o \neq$ adjacent node and $i \neq N$}{
    path$_{i}$=  ExplorePath(adjacent node, v$_o$, N)\;
      $i = i + 1$
    }
    \KwRet \{path$_i$\}\;
  }
  \Fn{\FMein{v$_s$, v$_o$, $N$}}
  {
    axiom\_path$_i$, adjacent node = GetNHopFromSource(v$_s$, 2) {\hfill\textbf{//}\,calculates 2 hop distance from source containing there exist and for all quantifier}\;
   
    axiom\_path$_i$ = hop\_path$_i$ $\cup$ axiom\_path$_{i-1}$ \;
    \If{$v_o \neq$ adjacent node and $i \neq N$}{
     axiom\_path$_i$=   ExploreSymbolicPath(adjacent node, v$_o$, N)\;
      $i = i + 1$
    } 
    \KwRet \{axiom\_path$_i$\} \;
  }
\end{algorithm}

\begin{figure}[ht]
\centering
\includegraphics[width=7cm]{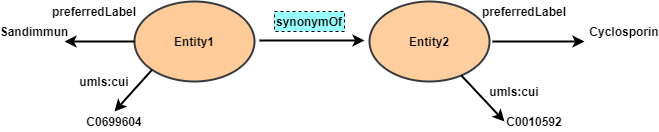}
\caption{{Subgraph of ontology illustrating direct connection between two entities}}
\label{single}
\end{figure}

\textbf{Single hop}. For retrieving the direct path, we query on the ontology using SPARQL to check if a path exists between entity pair (e$_s$, e$_o$). The study examined the potential interactions in a sentence \textit{\underline{Sandimmun}, a medication formulated as \underline{cyclosporin} (cya) in cremophor and ethanol, and the muscle relaxants atracurium and vecuronium in anesthetized cats.} The correct relation label between sandimmun and cyclosporin is \textit{hasTradename}. Upon querying this entity pair from the ontology, it was found that the direct path between given entity pairs is \textit{synonymOf} relation which is similar to the correct relation label \textit{hasTradename} present in the dataset. As depicted in Figure~\ref{single}, the direct path between two given entities (if they exist) is extricated using $\textrm{path}(y;e)\xrightarrow{}\ \textrm{cui}(x;y)\ \sqcap\ \textrm{edge}(x;z)\ \sqcap \textrm{cui}(z;e)$, where cui is concept unique identifier which uniquely identifies entity (assuming x is entity1, y is cui of entity1, z is entity2 and e is cui of entity2). It retrieves the connecting edge between two given entities. Once assimilating the path \textit{synonymOf} between entity pairs, the aggregator module in ReOnto computes the similarity between the extracted path and all relations, assigning the correct label \textit{hasTradename} as the similarity score reaches its maximum.
 
\begin{figure}[ht]
\centering
\includegraphics[width=8cm]{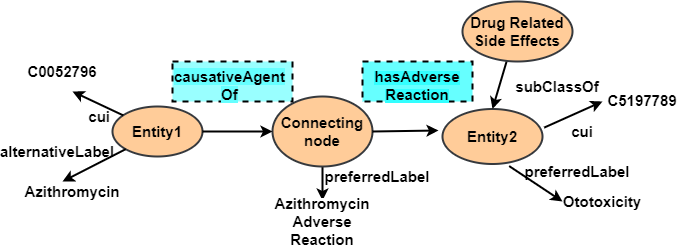}
\caption{{Subgraph of ontology depicting two hop distance between two entities}}
\label{path2}
\end{figure}

\textbf{Multi hop}. Multi-hop path reasoning over the knowledge base aims at finding a relation path for an entity pair by traversing along a path of triples from graph structure data \cite{lv2021multi}. For retrieving indirect path relation, we query on the ontology if a n-hop distance path exists between entity pair $(\textrm{e}_s,\textrm{e}_o)$ starting from 1-hop distance path. Consider the sentence \textit{Intravenous \underline{azithromycin}-induced \underline{ototoxicity}} with its relation label as \textit{hasAdverseEffect}. From ontologies,  we get the path as a concatenation of \textit{causative agent of, has adverse reaction} using \textrm{path}(y;e)$\xrightarrow{}$\ \textrm{cui}(x;y)$\ \sqcap$ \textrm{edge}(x;z)$\ \sqcap$\ \textrm{edge}(z;a) $\sqcap$ \textrm{cui}(a;e).
The aggregator module receives this path as input and using a similarity score, assigns the target relation label \textit{adverseEffect}. Refer to Figure~\ref{path2} for details.


\textbf{Using axioms}. So far, we have considered  only shallow and transitive relationships among the concepts. However, the biomedical domain consists of several complex relations. We argue that those relations can be captured using expressive axioms from ontology. Expressive axioms consist of logical quantifiers such as \textit{there exist$(\exists)$, for all$(\forall)$, union$(\sqcup)$, intersection$(\sqcap)$} which are part of popular biomedical ontologies. These expressive axioms enrich an ontology and play an essential role in the performance of downstream applications \cite{pan2019entity}. Our objective is to determine the relation label between two entities by tracing the corresponding multi-hop triplet path that contains these axioms, starting from the first entity in the graph and continuing up to a specified distance until we reach the second entity.
Note that when multiple paths are available between two entities, we have taken into account all the paths that are available which consist of unique keywords
Consider the sentence, \textit{A 73-year-old woman presented with fever and \underline{cough} 2 weeks after completing the third cycle of \underline{fludarabine} for chronic lymphocytic leukemia}. Here, correct relation label is \textit{adverse effect}. From the ontology, we get following sub-graph enriched with axioms.


  \begin{center}
       \small         
 {\textit{Fludaraline}} $\xrightarrow[]{\textit{causativeAgentOf}}$ \textit{{Fludaraline Adverse Reaction}} \\
  {\textit{Fludaraline Adverse Reaction}}${\sqsubseteq}
\exists${\textit{hasFinding.Finding}} \\  
 {\textit{Cough}\textit{$\sqsubseteq$}\textit{Finding}} \\

  \end{center}  


From the above relations, one can see that Fludaraline and Fludaraline Adverse Reaction (FADR) has a relation \textit{causativeAgentOf}. Moreover, there exists a \textit{hasFinding} relation between FADR and Finding. Therefore, with ontology reasoning, we can interpret that Fludaraline has an axiom path consisting of \textit{causativeAgentOf, hasSomeFinding}, which is closest to the relation label adverseEffect.
Similarly, consider another sentence, \textit{concentrations were significantly related to the degree of apocrine differentiation of the \underline{tumour} and, in a subset of the \underline{cancers}, capacity to release gcdfp-15 was positively correlated with incidence of progestogen and androgen receptors}. The labeled relation for this sentence is has \textit{nichdParentOf}
\begin{frame}

\small
\centering
  {{\textit{Tumor}$\xrightarrow[]{\textit{qualifierBy}}$\textit{Diagnostic\hspace{0.1cm}Imaging}} \\
{\textit{Diagnostic\hspace{0.1cm}Imaging}$\xrightarrow[]{\textit{allowedQualifier}}$\textit{Neoplasms}}
                {${\textit{Neoplasms} \sqsubseteq {\exists \textit{parent.Post-Traumatic Cancer}}}$} \\
                      {{\textit{Post-Traumatic Cancer} $\sqsubseteq$ \textit{Cancer}}}  \\
                   }  
\end{frame}
     

For the above case, the derived path is \textit{qualifierBy, allowedQualifier, subClass and there exist some parent and subClass}. 

\subsection{Encoding module}
Entity pairs are encoded by concatenating the position embedding with the word embedding in the sentence (Equation~\ref{eqn1}),  represented as $En(s_t^{s,o})$ where $s_t$ is the word embedding and $p_t^{s,o}$ is the position embedding at word position $t$ relative to the entity pair position $(s,o)$. Similarly, symbolic path information from the Symbolic Module ($SK$)  is encoded by concatenating path ($path_0^i$) and axiom path details $(axiom\_path_0^i)$ where $i$ represents the number of hops reaching destination.
\begin{equation} \label{eqn1}
    En(s_t^{s,o})=[s_t;p_t^{s,o}]
\end{equation}
\begin{equation} \label{enco}
    En(SK^{s,o})=[\sum path_0^i;\sum axiom\_path_0^i]^{s,o}
\end{equation}

The entity pairs representation  and path information, after encoding with BioBERT are forwarded to a  multi layer perceptron with non linear activation $\sigma$ (Equation \ref{eqn2} and \ref{eqn3}). We concatenate them as a shown in Equation~\ref{eqn4}. Since our dataset are from biomedical domain, we have used BioBERT for encoding. 


 \begin{equation} \label{eqn2}
 \small
A_{s,o}^n=MLP_n(BioBERT(En(s_0^{s,o}),En(s_1^{s,o}),..,En(s_l-1^{s,o})) 
\end{equation}
\begin{equation}\label{eqn3}
SP_{s,o}^n=MLP_n(BioBERT(En(SK^{s,o})))
\end{equation}
\begin{equation}\label{eqn4}
M_{s,o}^n=SP_{s,o}^{(n)}+A_{s,o}^{(n)}
\end{equation}

 \subsection{Graph Neural Network}
\subsubsection{Propagation module}
In this module, we propagate information among graph nodes using equation \ref{eqn5}, where given the representation of layer $n$ , representation of layer $n+1$ is calculated. Here $n$ represents the index, $B$ represents neighbors of $v_o$, and $\sigma$ is the nonlinear activation function.
 
 \begin{equation} \label{eqn5}
    h_s^{n+1}=\sum_{v_o\in B(v_o)}\sigma{({M_{s,o}^{(n)}h_o^{(n)}})}
\end{equation}

\subsubsection{Classification module}
 In the classification module, embeddings of entity pair are the input. Now, ReOnto performs element wise multiplication on input and then passed into multi layer perceptron using equation \ref{eqn6}. Here $\cdot$ represent element wise multiplication.
 
\begin{equation} \label{eqn6}
    MLP(v_s,v_o)=[h_{v_s}^{(1)} \cdot h_{v_o}^{(1)}]^T ; [h_{v_s}^{(2)} \cdot h_{v_o}^{(2)}]^T ;...;[h_{v_s}^{(K)} \cdot h_{v_o}^{(K)}]^T
\end{equation}

\subsubsection{Aggregator module}
Path information ($path_0^{i};axiom\_path_0^{i})$$\in$$ SK^{s,o}$ from Symbolic Module is separately encoded using BioBERT\footnote{https://www.sbert.net/} model, which is pretrained on biomedical text  corpora. At first, we perform encoding of path information and total relation label $R_1^i$ where $i$ is the total number of potential relations (refer Equation~\ref{eqn7} and~\ref{eqn8}). Then, we evaluate the semantic similarity between path information and complete labeled relation list. We get the relation label with the maximum similarity score and add it as a weighted bias as given in Equation~\ref{eqn9}. An important observation to make is that the weights generated by the GNN undergo modification by incorporating the knowledge of the Symbolic Module. This step is crucial as it involves combining the symbolic and sub-symbolic components. We employ the softmax function to obtain probabilities and compute the cross entropy loss (refer Equations~\ref{eqn10} and \ref{eqn11}), where S denotes whole corpus  and n are total entity pairs such that s $\neq$ o. It is worth noting that if no path exists between two entities, the bias score is set to 0, and loss is computed accordingly.
\begin{equation} \label{eqn7}
Renc=enc{(R_1^i)}
\end{equation}

\begin{equation} \label{eqn8}
    Penc=enc(SK^{s,o})
\end{equation}
\begin{equation} \label{eqn9}
    biasedscore_r=max{(cosSim{(Renc,Penc)})}
\end{equation}
\begin{equation} \label{eqn10}
    P(v_s,v_o)=softmax{((MLP{(v_{s},v_{o})}+biasedscore_r)}
\end{equation}

\begin{equation} \label{eqn11}
    L=\sum_{t=0}^{S}\sum_{s,o=0}^n (log P(v_s,v_o))_t
\end{equation}


\section{Experimental Setup}
We conduct our evaluation in response to following research questions.

\textbf{RQ1}: What is the effectiveness of ReOnto that combines symbolic knowledge with a neural model in solving biomedical relation extraction task?

\textbf{RQ2}: How does knowledge encoded in different ontologies impact performance of ReOnto? 

\textbf{Datasets}. Our initial biomedical dataset is BioRel~\cite{xing_BioRel_2020}, which includes a total of 533,560 sentences, 69,513 entities, and 125 relations. The second dataset we use is the Adverse Drug Effect (ADE) dataset~\cite{GURULINGAPPA2012885}. We treat the RE problem in this dataset as binary classification, where sentences are categorized as either positive adverse-related or negative adverse-related. Positive adverse relations are established when drug and reaction entities are associated in the given context, while negative relations involve drugs that are not accountable for a specific reaction. The ADE dataset comprises 6,821 labeled adverse sentences and 16,695 labeled negative adverse sentences, with a total of 5,063 entities. We consider two types of relations in this dataset: adverse-related and not adverse-related. The first entity is viewed as the drug, while the second entity is retrieved using named entity recognition. Table~\ref{ontology} provides details of the public ontologies utilized for constructing symbolic knowledge.

\begin{table}

    \centering
      \caption{{Ontologies used for Symbolic Knowledge}}

\label{ontology}
        \scalebox{0.8}{

    \begin{tabular}{ *{4}{c} }
      \toprule

      \makecell{Ontology} & \makecell{Classes} & \makecell{Properties} & \makecell{Maximum depth} \\
      \midrule
     DINTO~\cite{bona_enhancing_2019} &28,178 & 12& 2 \\
 OAE~\cite{he_oae_2014} & 10,589 & 123 & 17  \\
NDF-RT~\cite{winnenburg_using_2015} & 36,202 &  90 & 9 \\ 
MEDLINE~\cite{yang_ontology-based_2003} & 2,254 & 12 & 2 \\ 
NCIt~\cite{10.1007/11527770_30} & 177,762 & 97 & 21 \\ 
      \bottomrule
    \end{tabular}
    }  
\end{table}

\begin{table}[ht]
    \centering
     \begin{tabular}{ *{2}{c} }
      \toprule
      \makecell{Hyper-parameters } & \makecell{Value}  \\
      \midrule
     learning rate & 0.001  \\
    batch size & 50  \\
    dropout ratio & 0.5  \\
    hidden state size & 256  \\
    non linear activation & relu  \\
      \bottomrule
    \end{tabular}
    \caption{Hyper parameter seetings}
    \label{hyper}
\end{table}




\textbf{Baseline Models for comparison}. We used several competitive baselines: 1) Multi-instance models such as \cite{Nguyen2015RelationEP,zeng-etal-2014-relation,zeng-etal-2015-distant}, 2) Sentential RE models such as \cite{bastos2022how,zhu-etal-2019-graph,sorokin-gurevych-2017-context}. For Recon \cite{bastos2022how}, we used its EAC variant for fair comparison. Please note, we adapted these models to biomedical domain by re-training and inducing biomedical context needed for these models such as entity descriptions and types. 3) Biomedical relation extraction works such as \cite{Huynh2016AdverseDR,electronics11203336,haq_mining_2022,schlichtkrull2017modeling,xing_BioRel_2020}. For biomedical RE works, values are obtained from original papers, and for other works (sentential and multi-instance), if code is available, we executed them on both datasets. 

\textbf{Hyper-parameters and Metrics}. Table~\ref{hyper} outlines the best parameter setting. We employ GloVe embedding of dimension 50 for initialization. Since the datasets are from the biomedical domain for evaluating semantic similarity, we have used BioBERT model\footnote{https://www.sbert.net/}. The size of position embedding is also kept at 50. 
We have used the open-source ontology (.owl) from BioPortal to extract the paths using the SPARQL query.  We have followed \cite{zhu-etal-2019-graph} for experiment settings. We evaluated the accuracy (precision) and F1 score for both datasets.


\begin{table}[!htb]
  \caption{{Biomedical Relation Extraction Results. ReOnto outperforms baselines on both datasets. We've left precision column blank for baselines that does not report it. }}\label{results}  
    \begin{center}
        \subtable{
        \scalebox{0.7}{
    \begin{tabular}{ccccc}
\hline
{Dataset} & {Model} & {Accuracy(in\%)} & {F1 scores} \\
\hline

\multirow{3}{4em}{ADE} & CNN~\cite{Nguyen2015RelationEP} & 68 & 0.71  \\ 
& PCNN~\cite{zeng-etal-2015-distant} & 76.9 & 0.73\\ 
&ContextAware~\cite{sorokin-gurevych-2017-context} &93 & 0.93\\ 
&RGCN~\cite{schlichtkrull2017modeling} &86 & 0.83\\ 

&GPGNN~\cite{zhu-etal-2019-graph} &92.1 & 0.90\\ 
&CRNN~\cite{Huynh2016AdverseDR} & - & 0.87\\ 
&CNN-Embedding~\cite{electronics11203336} & - & 0.89\\ 

&SparkNLP~\cite{haq_mining_2022} & - & 0.85\\ 
&T5~\cite{raffel2020exploring} & 92 & 0.86\\ 

&RECON~\cite{bastos2021recon} & 93.5 & 0.92\\ 

&ReOnto \textbf{(Ours)}& \textbf{97} & \textbf{0.96}\\ 
\hline

            \end{tabular}
            }
            \centering
        } 
        \scalebox{0.7}{
        \subtable{
          \centering
            \begin{tabular}{ccccc}
\hline
{Dataset} & {Model} & {Accuracy(in\%)} & {F1 scores} \\
\hline
\multirow{3}{4em}{BioRel} & CNN~\cite{Nguyen2015RelationEP} & 48 & 0.47  \\ 
& PCNN~\cite{zeng-etal-2015-distant} & 64.6 & 0.57\\ 
&RGCN~\cite{schlichtkrull2017modeling} &72 & 0.78\\ 
&GPGNN~\cite{zhu-etal-2019-graph} &85 & 0.84\\ 
&CNN+ATT~\cite{xing_BioRel_2020} & - &0.72 \\
&PCNN+AVG~\cite{xing_BioRel_2020}& - & 0.76 \\
&RNN+AVG~\cite{xing_BioRel_2020} & - & 0.74 \\
&ContextAware~\cite{sorokin-gurevych-2017-context} & 89 & 0.87\\ 
&T5~\cite{raffel2020exploring} & 88 & 0.86 \\ 

&RECON~\cite{bastos2021recon} & 89.6 & 0.86\\ 
&ReOnto \textbf{(Ours)} & \textbf{92} & \textbf{0.90}\\ 
\hline
\end{tabular}  

}
        } 
    \end{center}
  

\end{table}

\section{Results}
\begin{figure*}[ht]
    \centering
    \includegraphics[scale=0.20]{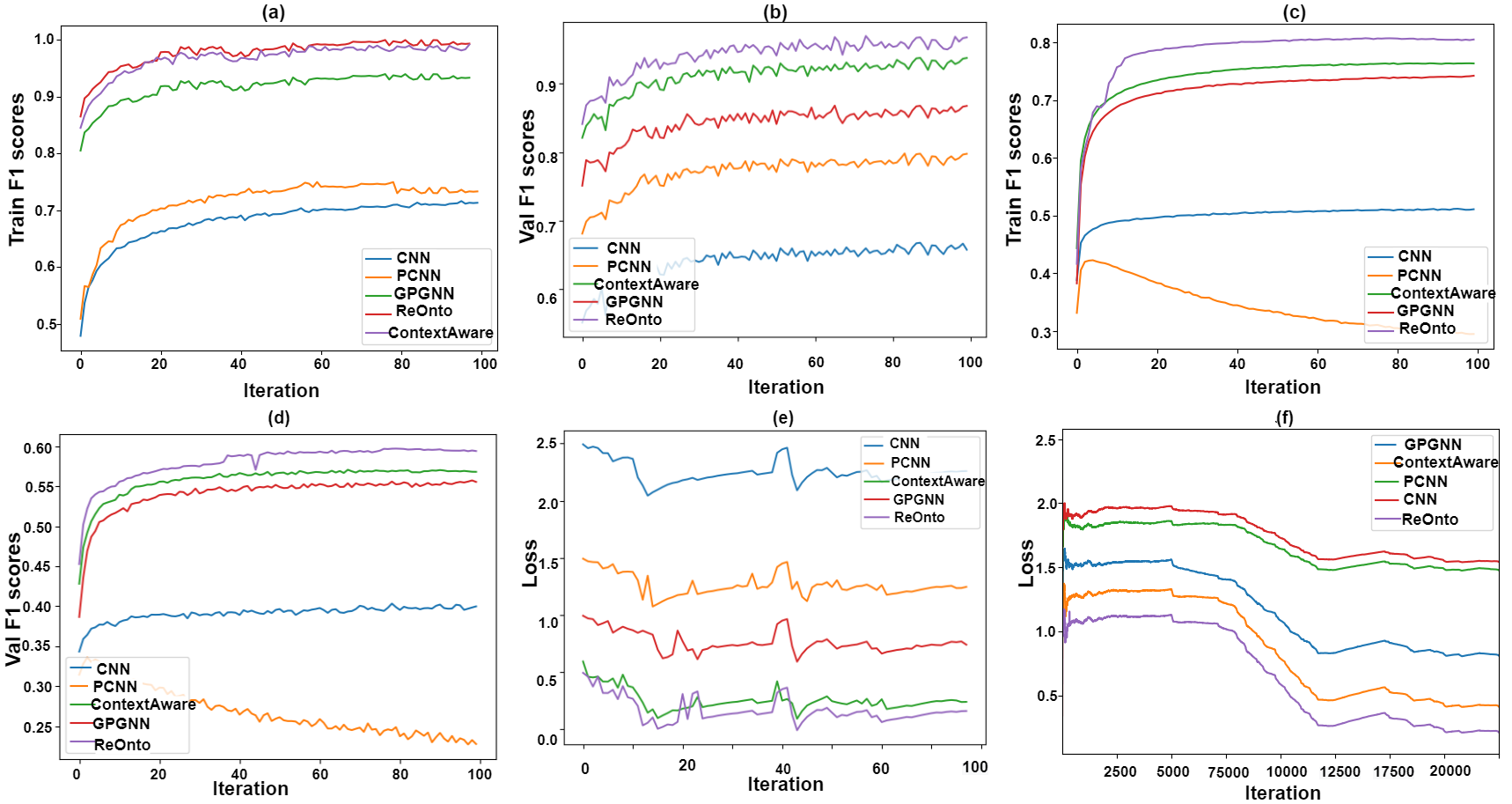}
    \caption{{For the ADE dataset, Figures a) and b) show the training and validation F1 scores with baseline, respectively. Figure e) illustrates the cross-entropy loss for the iteration. For the BioRel dataset, Figures c) and d) show the training and validation F1 scores with baseline, respectively. Figure f) illustrates the cross-entropy loss concerning the iteration. ReOnto exhibits consistent and stable performance on both datasets, as indicated by the plotted F1 scores and loss.
}}
    \label{res}
\end{figure*}

ReOnto outperforms all the baseline models on both datasets (From Table~\ref{results}). These results indicate that our model could successfully conduct reasoning with a neuro-symbolic graph on the fully connected graph and combine it with the underlying deep learning model (GNN in our case). Observed results successfully answer \textbf{RQ1}. Methods such as \cite{bastos2021recon,sorokin-gurevych-2017-context} use contexts such as entity types and descriptions. Similarly, RECON and T5 include additional explicit information of long entity descriptions, its type that allows offline learning of entity context. However, in a real-world setting of the biomedical domain, it is viable that such context may not be present for each entity. In contrast, our model discards the necessity of available entity context and learns purely using reasoning over connected entity graphs. Furthermore, multi-instance baselines try to learn relations using previous occurrences of entities in the document. In both cases, missing reasoning to capture long-range dependencies of entities hampers their performance. 
One possible reason for CNN and PCNN not performing well is that the biomedical sentence is complex and direct adherence to relation is impossible in this type of text. We can also notice that the context-aware model is performing better than multi-instance on these datasets because entity contexts are helping up to an extent. Presently, we have added context information(symbolic knowledge) via ontology into the model. If enough context details are given our model can work on generalised datasets as well.
Figure~\ref{res} presents plots a, b, c, d, which depict the training and validation F1 scores on both datasets, while plots e, f show the loss graph. Our observations indicate that ReOnto delivers consistent performance on these graphs within the considered timeframe.

\section{Ablation study}

\subsection{Effectiveness of number of ontologies}
To better understand the contribution of each ontology on ReOnto's performance, we conducted an ablation study. Table~\ref{tab:ontologyperformance} presents a summary of our findings, which indicate a significant decrease in performance when considering individual ontologies. This validates our approach of merging knowledge from multiple ontologies to create symbolic knowledge.

For the ADE dataset, we have a lesser entity coverage of 22\% using DRON ontology. However, we found that the performance significantly improves when we increase the entity coverage by incorporating the OAE and DINTO ontologies. This increase in entity coverage results in corresponding improvements in F1 scores. Similarly, for the BioRel dataset, we tested with MEDLINE ontology with entity coverage of 42\% and then NCIt ontology with coverage of 34\%, leading to corresponding improvements in F1 scores. Results also provide conclusive evidence that ReOnto's performance depends on the coverage of entities aligned with the dataset and combining encoded knowledge has positive impact on overall performance (answering \textbf{RQ2}).


\begin{table}[ht] 
\small
\centering
 \caption{{Effect of ontology on F1 scores}}
    \label{tab:ontologyperformance}   

\scalebox{0.70}{

\begin{tabular}{cccc}
\hline

{Dataset} & {Ontology} & {Entity coverage(approx.)} & {F1 scores} \\
\hline
\multirow{3}{4em}{ADE} & DRON~\cite{bona_enhancing_2019} & 22\% & 0.92  \\ 
& OAE~\cite{he_oae_2014} & 34\% & 0.93\\ 
& DINTO~\cite{herrero-zazo_dinto_2015} &41\% & 0.95\\ 
\hline
\multirow{3}{4em}{BioRel} & MEDLINE~\cite{yang_ontology-based_2003} & 42\% & 0.88 \\ 
& NCIt~\cite{10.1007/11527770_30} & 34\% & 0.84 \\ 
\hline
\end{tabular}
}
\end{table}


\subsection{Effectiveness of number of hops}
\begin{figure}[ht]
    
  \centering
  \includegraphics[scale=0.30]{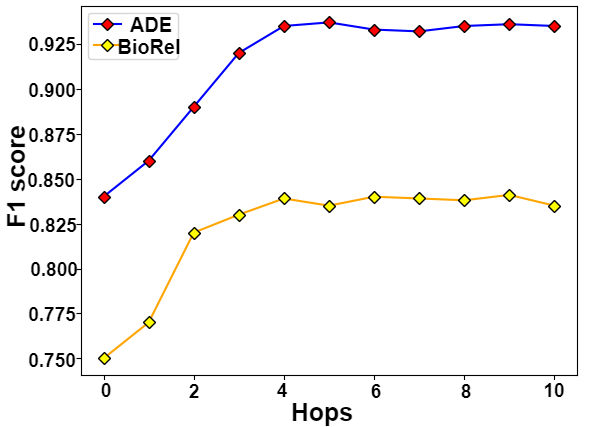}
\caption{{Effectiveness of hops on performance}}
    \label{hop}
\end{figure}

We separately study the effect of the number of hops on the performance of ReOnto. 
Figure~\ref{hop} shows the impact of the number of hops on the model. Increasing hops initially improve F1 scores until reaching a plateau. This is because additional hops don't provide new relevant information. Table~\ref{hoptable} summarizes the extracted hops from the MEDLINE ontology, supporting our observation. Interestingly, increasing hops leads to redundant information that doesn't contribute to performance. To maintain context and meaningful connections, we preserved multi-hop information up to five hops in our experiment. Furthermore, Table~\ref{parsingtime} illustrates the relationship between ontology size, parsing time, and the number of hops, indicating an increase in time as hops increase.

\begin{table}[ht] 
\small
\centering
    \caption{{Time taken to parse ontology and evaluate respective path. Parsing time increase w.r.t size of ontology}}\label{parsingtime}
\scalebox{0.65}{
\begin{tabular}{cccccccc}

\hline
 Ontology& Size (in KB)&  \multicolumn{6}{c}{\text{Time taken (in seconds)}} \\
\hline

 & & Parsing & Direct hop & One hop & Two hop & Axiom path1  & Axiom path2\\ 
 \hline

OAE & 9286 & 6.75 &   0.11 & 2.73  & 7.47 & 2.19 & 5.92 \\ 
NDFRT & 69387 & 123.11 & 1.21 & 0.003 & 7.629 & 44.79 & 103.36  \\
DINTO & 1,10,865 & 137.4 & 1.6 & 3.8  & 8.54 & 5.67 & 11.32 \\
MEDLINE & 6975 & 2.19 & 0.002 &  0.0023 & 0.003 & 3.118
 & 6.09  \\ 
NCI & 5,71,434 & 758.9 &  1034.5 & 1294.5 &  3454.1 & 2485 & 5569 \\ 
    \hline

\end{tabular}}

\end{table}

\begin{table}[ht]
        \centering
    \caption{{Derived path obtained by connecting  ``protein" and  ``dietary protein" entity}}
    \label{hoptable}
    \scalebox{0.6}{

   \begin{tabular}{p{1cm} p{13cm} }
    \hline
    {Hops} & {\hspace{3cm}Path} \\
    \hline
    path1 & classifies \\
 path2 & mapped from  dietary proteins, classifies  \\
 path3 & classifies proteins,  classifies dietary proteins, classifies \\
 path4 & classifies proteins, classifies dietary proteins, mapped from dietary proteins, classifies \\ 
 path5 & classifies proteins, classifies dietary proteins, related to carbs, related to dietary proteins, classifies \\
 path6 & classifies Proteins, classifies dietary proteins, mapped from dietary proteins, related to carbs, related to dietary proteins, classifies \\ 
 path7 & classifies Proteins, classifies dietary proteins, mapped from dietary proteins, related to carbs, related to dietary proteins classifies dietary proteins\\ 
    \hline
    \end{tabular}
}


\end{table}


\begin{table*}[ht]
\centering

\caption{{Sample sentences and predictions of various models. ReOnto using reasoning is able to predict the relations which are not explicitly observable from the sentence itself and requires long-range entity interactions.}}\label{casestudy}
\scalebox{0.8}{
\begin{tabular}{p{8cm}@{\hspace{0.5cm}}p{1.5cm}@{\hspace{0.5cm}}p{1.5cm}@{\hspace{0.5cm}}p{1.5cm}@{\hspace{0.5cm}}p{1.5cm}}
\hline
\textbf{Sentence}
& \textbf{Relation} & \textbf{GPGNN} & \textbf{Context Aware} & \textbf{ReOnto} \\ \hline

 Both compounds are equally potent in the stimulation of intestinal calcium transport , \underline{bone}(C0262950) calcium mobilization , in the elevation of serum phosphorus , and in the healing of \underline{rickets}(C0035579) in the rat
& is primary anatomic site of disease  & may be associated disease of disease & may be finding of disease & is primary anatomic site of disease \\
\hline The ventricular effective refractory period, as well as the \underline{vt cycle
length}(C0042514), increased after \underline{propranolol}(C0033497) and was further prolonged after
the addition of a type i agent & may be treated by & may diagnose & may treat & may be treated by \\ 
\hline
\hline dsip and clip [acth(18-39)] immunoreactive(ir) neurons and fibers were examined in the \underline{human}(C0086418) hypophysis and pituitary stalk using immunmohistofluorescence and \underline{peroxidase}(C4522012) antiperoxidase methods & is organism source of gene product  & nichd parent of
 & organism has gene &  is organism source of gene product   \\ 

\hline

\end{tabular}

}
\end{table*}
\section{Case study}
Table~\ref{casestudy} shows qualitative results that compare the ReOnto model with the baseline models. We report a few results showing ReOnto can surmise the relationship with reasoning.
ReOnto retrieved the relevant derived path from the ontology in the first case. ReOnto implicitly learns from the facts and captures the derived path to provide the correct relation label, even if it is not explicitly mentioned as  \textit{isPrimaryAnatomicSiteOfDisease}. 
\begin{tcolorbox}
\small
\textit{CUI:C0262950}$\xrightarrow[]{\textit{preferredLabel}}$\textit{Bone} \\
\textit{Bone}${\sqsubseteq{\exists\textit{anatomicSiteOfDisease.}}}$\textit{Rickets}  \\
\textit{Rickets}$\xrightarrow[]{\textit{CUI}}$\textit{CUI:C0035579} \\
\textit{Bone}$\xrightarrow[]{\textit{semanticType}}$\textit{Anatomic Structure}
\end{tcolorbox}
In the second case, ReOnto produces the following path by utilizing the expressive axiom of the ontology. ReOnto captures the long-range dependencies between entities and provides the correct relation label.

\begin{tcolorbox}
\small

\textit{CUI:C0042514}$\xrightarrow[]
{\textit{preferredLabel}}$\textit{Ventricular Tachycardia}

\textit{Ventricular\hspace{0.1cm}Tachycardia}$\equiv${\textit{Techycardia}} \\
\textit{Techycardia}${\sqsubseteq{\exists\textit{mayBeTreatedBy.Propranolol}}}$ \\
\textit{Propranolol}$\xrightarrow[]{\textit{CUI}}$  
\textit{CUI:C0033497}
\end{tcolorbox}

In the last case study, several paths are derived from the ontology for the \textit{Human} entity. It can be observed that ReOnto derives and asserts the dependency path between \textit{Human} and \textit{Peroxidase}, and concludes that the target relation label \textit{isOrganismSourceOfGeneProduct} applies, as compared to other baseline models. Such complex ontology reasoning provides long-range interactions between entities, which is inherently not possible in baseline models.

\begin{tcolorbox}
\small
\textit{CUI:C0086418}$\xrightarrow[]{\textit{preferredLabel}}$\textit{Human} 

$\exists$\textit{Human}$\sqsubseteq${\textit{geneProductHasOrganismSource.}}\textit{Myeloperoxidase} 

 \textit{Myeloperoxidase}$\xrightarrow[]{\textit{hasDisposition}}$\textit{Peroxidase}\textit{(disposition)} \\
\textit{Peroxidase(disposition)}$\xrightarrow[]{\textit{preferredName}}$\textit{Peroxidase}  
  \hspace{2cm}

\textit{Peroxidase}$\xrightarrow[]{\textit{CUI}}$ \textit{CUI:C4522012}

\end{tcolorbox}

\section{Conclusion and Future Work}
We proposed a novel neuro-symbolic approach ReOnto that leverages path-based reasoning, including expressive axiom path with GNN. We apply our model to complex biomedical text and compare the approach with baselines. With empirical results, there are three key takeaways. Firstly, existing baseline models with any form of context only capture short-range dependencies of entities. In contrast, our model uses long-range entity dependencies derived from ontology reasoning to outperform all baselines on both biomedical datasets. Code is available at \url{https://github.com/kracr/reonto-relation-extraction}. 

ReOnto provides effective reasoning on given text and entity pair, which can tackle the challenges of biomedical text. It also considers expressive axioms of ontology to reason on RE. The aggregation of these axioms outperformed the baselines. As a next step, we can consider using background knowlege on unsupervised data. An ontology reasoner can be used to infer more paths and perhaps these additional axioms can improve the performance further.

\textbf{Acknowledgements}. We express our sincere gratitude to the Infosys Centre for Artificial Intelligence (CAI) at IIIT-Delhi for their  support throughout the duration of this project.

\bibliographystyle{named}
\bibliography{ijcai23}

\begin{thebibliography}{}

\bibitem[\protect\citeauthoryear{Aghaebrahimian \bgroup \em et al.\egroup
  }{2022}]{aghaebrahimian2022ontology}
Ahmad Aghaebrahimian, Maria Anisimova, and Manuel Gil.
\newblock Ontology-aware biomedical relation extraction.
\newblock {\em bioRxiv}, 2022.

\bibitem[\protect\citeauthoryear{Angell \bgroup \em et al.\egroup
  }{2021}]{angell2021clustering}
Rico Angell, Nicholas Monath, Sunil Mohan, Nishant Yadav, and Andrew McCallum.
\newblock Clustering-based inference for biomedical entity linking.
\newblock In {\em Proceedings of the 2021 Conference of the North American
  Chapter of the Association for Computational Linguistics: Human Language
  Technologies}, pages 2598--2608, 2021.

\bibitem[\protect\citeauthoryear{Balcilar \bgroup \em et al.\egroup
  }{2020}]{balcilar2020analyzing}
Muhammet Balcilar, Guillaume Renton, Pierre H{\'e}roux, Benoit Ga{\"u}z{\`e}re,
  S{\'e}bastien Adam, and Paul Honeine.
\newblock Analyzing the expressive power of graph neural networks in a spectral
  perspective.
\newblock In {\em International Conference on Learning Representations}, 2020.

\bibitem[\protect\citeauthoryear{Bastos \bgroup \em et al.\egroup
  }{2021}]{bastos2021recon}
Anson Bastos, Abhishek Nadgeri, Kuldeep Singh, Isaiah~Onando Mulang, Saeedeh
  Shekarpour, Johannes Hoffart, and Manohar Kaul.
\newblock Recon: relation extraction using knowledge graph context in a graph
  neural network.
\newblock In {\em Proceedings of the Web Conference 2021}, pages 1673--1685,
  2021.

\bibitem[\protect\citeauthoryear{Bastos \bgroup \em et al.\egroup
  }{2022}]{bastos2022how}
Anson Bastos, Abhishek Nadgeri, Kuldeep Singh, Hiroki Kanezashi, Toyotaro
  Suzumura, and Isaiah~Onando Mulang'.
\newblock How expressive are transformers in spectral domain for graphs?
\newblock {\em Transactions on Machine Learning Research}, 2022.

\bibitem[\protect\citeauthoryear{Bodenreider}{2004}]{bodenreider2004unified}
Olivier Bodenreider.
\newblock The unified medical language system (umls): integrating biomedical
  terminology.
\newblock {\em Nucleic acids research}, 32(suppl\_1):D267--D270, 2004.

\bibitem[\protect\citeauthoryear{Bona \bgroup \em et al.\egroup
  }{2019}]{bona_enhancing_2019}
Jonathan~P. Bona, Mathias Brochhausen, and William~R. Hogan.
\newblock Enhancing the drug ontology with semantically-rich representations of
  national drug codes and {RxNorm} unique concept identifiers.
\newblock {\em BMC Bioinformatics}, 20(21), December 2019.

\bibitem[\protect\citeauthoryear{Cabot and Navigli}{2021}]{cabot2021rebel}
Pere-Llu{\'\i}s~Huguet Cabot and Roberto Navigli.
\newblock Rebel: Relation extraction by end-to-end language generation.
\newblock In {\em Findings of the Association for Computational Linguistics:
  EMNLP 2021}, pages 2370--2381, 2021.

\bibitem[\protect\citeauthoryear{Crone}{2020}]{crone2020deeper}
Phil Crone.
\newblock Deeper task-specificity improves joint entity and relation
  extraction.
\newblock {\em arXiv preprint arXiv:2002.06424}, 2020.

\bibitem[\protect\citeauthoryear{Eberts and Ulges}{2020}]{eberts2020span}
Markus Eberts and Adrian Ulges.
\newblock Span-based joint entity and relation extraction with transformer
  pre-training.
\newblock In {\em ECAI 2020}, pages 2006--2013. IOS Press, 2020.

\bibitem[\protect\citeauthoryear{Fensel \bgroup \em et al.\egroup
  }{2020}]{fensel2020we}
Dieter Fensel, Umutcan {\c{S}}im{\c{s}}ek, Kevin Angele, Elwin Huaman, Elias
  K{\"a}rle, Oleksandra Panasiuk, Ioan Toma, J{\"u}rgen Umbrich, and Alexander
  Wahler.
\newblock Why we need knowledge graphs: Applications.
\newblock In {\em Knowledge Graphs}, pages 95--112. Springer, 2020.

\bibitem[\protect\citeauthoryear{Gurulingappa \bgroup \em et al.\egroup
  }{2012}]{GURULINGAPPA2012885}
Harsha Gurulingappa, Abdul~Mateen Rajput, Angus Roberts, Juliane Fluck, Martin
  Hofmann-Apitius, and Luca Toldo.
\newblock Development of a benchmark corpus to support the automatic extraction
  of drug-related adverse effects from medical case reports.
\newblock {\em Journal of Biomedical Informatics}, 45(5):885 -- 892, 2012.
\newblock Text Mining and Natural Language Processing in Pharmacogenomics.

\bibitem[\protect\citeauthoryear{Haq \bgroup \em et al.\egroup
  }{2022}]{haq_mining_2022}
Hasham~Ul Haq, Veysel Kocaman, and David Talby.
\newblock Mining adverse drug reactions from unstructured mediums at scale,
  2022.
\newblock version: 2.

\bibitem[\protect\citeauthoryear{He \bgroup \em et al.\egroup
  }{2014}]{he_oae_2014}
Yongqun He, Sirarat Sarntivijai, Yu~Lin, Zuoshuang Xiang, Abra Guo, Shelley
  Zhang, Desikan Jagannathan, Luca Toldo, Cui Tao, and Barry Smith.
\newblock {OAE}: The ontology of adverse events.
\newblock {\em Journal of Biomedical Semantics}, 5:29, dec 2014.

\bibitem[\protect\citeauthoryear{Herrero-Zazo \bgroup \em et al.\egroup
  }{2015}]{herrero-zazo_dinto_2015}
María Herrero-Zazo, Isabel Segura-Bedmar, Janna Hastings, and Paloma
  Martínez.
\newblock {DINTO}: Using {OWL} ontologies and {SWRL} rules to infer drug–drug
  interactions and their mechanisms.
\newblock {\em Journal of Chemical Information and Modeling}, 55(8):1698--1707,
  aug 2015.
\newblock Publisher: American Chemical Society.

\bibitem[\protect\citeauthoryear{Hong \bgroup \em et al.\egroup
  }{2004}]{hong_ontology-based_2004}
Jia-Fei Hong, Xiang-Bing Li, and Chu-Ren Huang.
\newblock Ontology-based prediction of compound relations : A study based on
  {SUMO}.
\newblock In {\em Proceedings of the 18th Pacific Asia Conference on Language,
  Information and Computation}, pages 151--160. Logico-Linguistic Society of
  Japan, dec 2004.

\bibitem[\protect\citeauthoryear{Huynh \bgroup \em et al.\egroup
  }{2016}]{Huynh2016AdverseDR}
Trung-Tin Huynh, Yulan He, Alistair Willis, and Stefan~M. R{\"u}ger.
\newblock Adverse drug reaction classification with deep neural networks.
\newblock In {\em International Conference on Computational Linguistics}, 2016.

\bibitem[\protect\citeauthoryear{Karkaletsis \bgroup \em et al.\egroup
  }{2011}]{karkaletsis_ontology_2011}
Vangelis Karkaletsis, Pavlina Fragkou, Georgios Petasis, and Elias Iosif.
\newblock Ontology based information extraction from text.
\newblock In Georgios Paliouras, Constantine~D. Spyropoulos, and George
  Tsatsaronis, editors, {\em Knowledge-Driven Multimedia Information Extraction
  and Ontology Evolution: Bridging the Semantic Gap}, pages 89--109. Springer
  Berlin Heidelberg, 2011.

\bibitem[\protect\citeauthoryear{Kumar and Smith}{2005}]{10.1007/11527770_30}
Anand Kumar and Barry Smith.
\newblock Oncology ontology in the nci thesaurus.
\newblock In Silvia Miksch, Jim Hunter, and Elpida~T. Keravnou, editors, {\em
  Artificial Intelligence in Medicine}, pages 213--220, Berlin, Heidelberg,
  2005. Springer Berlin Heidelberg.

\bibitem[\protect\citeauthoryear{Li and Huan}{2008}]{ontology}
D.~Li and L.~Huan.
\newblock The ontology relation extraction for semantic web annotation.
\newblock In {\em Cluster Computing and the Grid, IEEE International Symposium
  on}, pages 534--541, Los Alamitos, CA, USA, may 2008. IEEE Computer Society.

\bibitem[\protect\citeauthoryear{Lv \bgroup \em et al.\egroup
  }{2021}]{lv2021multi}
Xin Lv, Yixin Cao, Lei Hou, Juanzi Li, Zhiyuan Liu, Yichi Zhang, and Zelin Dai.
\newblock Is multi-hop reasoning really explainable? towards benchmarking
  reasoning interpretability.
\newblock In {\em Proceedings of the 2021 Conference on Empirical Methods in
  Natural Language Processing}, pages 8899--8911, 2021.

\bibitem[\protect\citeauthoryear{Nadgeri \bgroup \em et al.\egroup
  }{2021}]{nadgeri-etal-2021-kgpool}
Abhishek Nadgeri, Anson Bastos, Kuldeep Singh, Isaiah~Onando Mulang{'},
  Johannes Hoffart, Saeedeh Shekarpour, and Vijay Saraswat.
\newblock {KGP}ool: Dynamic knowledge graph context selection for relation
  extraction.
\newblock In {\em Findings of the Association for Computational Linguistics:
  ACL-IJCNLP 2021}, pages 535--548, Online, August 2021. Association for
  Computational Linguistics.

\bibitem[\protect\citeauthoryear{Nguyen and
  Grishman}{2015}]{Nguyen2015RelationEP}
Thien~Huu Nguyen and Ralph Grishman.
\newblock Relation extraction: Perspective from convolutional neural networks.
\newblock In {\em VS@HLT-NAACL}, 2015.

\bibitem[\protect\citeauthoryear{Pan \bgroup \em et al.\egroup
  }{2019}]{pan2019entity}
Jeff~Z Pan, Mei Zhang, Kuldeep Singh, Frank~van Harmelen, Jinguang Gu, and Zhi
  Zhang.
\newblock Entity enabled relation linking.
\newblock In {\em International Semantic Web Conference}, pages 523--538.
  Springer, 2019.

\bibitem[\protect\citeauthoryear{Raffel \bgroup \em et al.\egroup
  }{2020}]{raffel2020exploring}
Colin Raffel, Noam Shazeer, Adam Roberts, Katherine Lee, Sharan Narang, Michael
  Matena, Yanqi Zhou, Wei Li, and Peter~J. Liu.
\newblock Exploring the limits of transfer learning with a unified text-to-text
  transformer, 2020.

\bibitem[\protect\citeauthoryear{Rawat \bgroup \em et al.\egroup
  }{2022}]{electronics11203336}
Ashish Rawat, Mudasir~Ahmad Wani, Mohammed ElAffendi, Ali~Shariq Imran, Zenun
  Kastrati, and Sher~Muhammad Daudpota.
\newblock Drug adverse event detection using text-based convolutional neural
  networks (textcnn) technique.
\newblock {\em Electronics}, 11(20), 2022.

\bibitem[\protect\citeauthoryear{Santos \bgroup \em et al.\egroup
  }{2015}]{https://doi.org/10.48550/arxiv.1504.06580}
Cicero Nogueira~dos Santos, Bing Xiang, and Bowen Zhou.
\newblock Classifying relations by ranking with convolutional neural networks,
  2015.

\bibitem[\protect\citeauthoryear{Schlichtkrull \bgroup \em et al.\egroup
  }{2017}]{schlichtkrull2017modeling}
Michael Schlichtkrull, Thomas~N Kipf, Peter Bloem, Rianne van~den Berg, Ivan
  Titov, and Max Welling.
\newblock Modeling relational data with graph convolutional networks.
\newblock {\em arXiv preprint arXiv:1703.06103}, 2017.

\bibitem[\protect\citeauthoryear{Shen and
  Huang}{2016}]{shen-huang-2016-attention}
Yatian Shen and Xuanjing Huang.
\newblock Attention-based convolutional neural network for semantic relation
  extraction.
\newblock In {\em Proceedings of {COLING} 2016, the 26th International
  Conference on Computational Linguistics: Technical Papers}, pages 2526--2536,
  Osaka, Japan, December 2016. The COLING 2016 Organizing Committee.

\bibitem[\protect\citeauthoryear{Sorokin and
  Gurevych}{2017}]{sorokin-gurevych-2017-context}
Daniil Sorokin and Iryna Gurevych.
\newblock Context-aware representations for knowledge base relation extraction.
\newblock In {\em Proceedings of the 2017 Conference on Empirical Methods in
  Natural Language Processing}, pages 1784--1789, Copenhagen, Denmark,
  September 2017. Association for Computational Linguistics.

\bibitem[\protect\citeauthoryear{T.Y.S.S \bgroup \em et al.\egroup
  }{2021}]{santosh2021joint}
Santosh T.Y.S.S, Prantika Chakraborty, Sudakshina Dutta, Debarshi~Kumar Sanyal,
  and Partha~Pratim Das.
\newblock Joint entity and relation extraction from scientific documents: Role
  of linguistic information and entity types.
\newblock In {\em EEKE@JCDL}, 2021.

\bibitem[\protect\citeauthoryear{Vashishth \bgroup \em et al.\egroup
  }{2018}]{vashishth-etal-2018-reside}
Shikhar Vashishth, Rishabh Joshi, Sai~Suman Prayaga, Chiranjib Bhattacharyya,
  and Partha Talukdar.
\newblock {RESIDE}: Improving distantly-supervised neural relation extraction
  using side information.
\newblock In {\em Proceedings of the 2018 Conference on Empirical Methods in
  Natural Language Processing}, pages 1257--1266, Brussels, Belgium,
  October-November 2018. Association for Computational Linguistics.

\bibitem[\protect\citeauthoryear{Wang \bgroup \em et al.\egroup
  }{2016}]{wang-etal-2016-relation}
Linlin Wang, Zhu Cao, Gerard de~Melo, and Zhiyuan Liu.
\newblock Relation classification via multi-level attention {CNN}s.
\newblock In {\em Proceedings of the 54th Annual Meeting of the Association for
  Computational Linguistics (Volume 1: Long Papers)}, pages 1298--1307, Berlin,
  Germany, August 2016. Association for Computational Linguistics.

\bibitem[\protect\citeauthoryear{Winnenburg \bgroup \em et al.\egroup
  }{2015}]{winnenburg_using_2015}
Rainer Winnenburg, Jonathan~M Mortensen, and Olivier Bodenreider.
\newblock Using description logics to evaluate the consistency of drug-class
  membership relations in {NDF}-{RT}.
\newblock {\em Journal of Biomedical Semantics}, 6:13, 2015.

\bibitem[\protect\citeauthoryear{Xing \bgroup \em et al.\egroup
  }{2020}]{xing_BioRel_2020}
Rui Xing, Jie Luo, and Tengwei Song.
\newblock {BioRel}: towards large-scale biomedical relation extraction.
\newblock {\em BMC Bioinformatics}, 21(16):543, 2020.

\bibitem[\protect\citeauthoryear{Yang}{2003}]{yang_ontology-based_2003}
Jung-Jin Yang.
\newblock An ontology-based intelligent agent system for semantic search in
  medicine.
\newblock In Jaeho Lee and Mike Barley, editors, {\em Intelligent Agents and
  Multi-Agent Systems}, pages 182--193. Springer Berlin Heidelberg, 2003.

\bibitem[\protect\citeauthoryear{Ye and
  Ling}{2019}]{https://doi.org/10.48550/arxiv.1904.00143}
Zhi-Xiu Ye and Zhen-Hua Ling.
\newblock Distant supervision relation extraction with intra-bag and inter-bag
  attentions, 2019.

\bibitem[\protect\citeauthoryear{Zeng \bgroup \em et al.\egroup
  }{2014}]{zeng-etal-2014-relation}
Daojian Zeng, Kang Liu, Siwei Lai, Guangyou Zhou, and Jun Zhao.
\newblock Relation classification via convolutional deep neural network.
\newblock In {\em Proceedings of {COLING} 2014, the 25th International
  Conference on Computational Linguistics: Technical Papers}, pages 2335--2344,
  Dublin, Ireland, August 2014. Dublin City University and Association for
  Computational Linguistics.

\bibitem[\protect\citeauthoryear{Zeng \bgroup \em et al.\egroup
  }{2015}]{zeng-etal-2015-distant}
Daojian Zeng, Kang Liu, Yubo Chen, and Jun Zhao.
\newblock Distant supervision for relation extraction via piecewise
  convolutional neural networks.
\newblock In {\em Proceedings of the 2015 Conference on Empirical Methods in
  Natural Language Processing}, pages 1753--1762, Lisbon, Portugal, September
  2015. Association for Computational Linguistics.

\bibitem[\protect\citeauthoryear{Zhang and
  Wang}{2015}]{https://doi.org/10.48550/arxiv.1508.01006}
Dongxu Zhang and Dong Wang.
\newblock Relation classification via recurrent neural network, 2015.

\bibitem[\protect\citeauthoryear{Zhang \bgroup \em et al.\egroup
  }{2022}]{zhang2022knowledge}
Wen Zhang, Jiaoyan Chen, Juan Li, Zezhong Xu, Jeff~Z Pan, and Huajun Chen.
\newblock Knowledge graph reasoning with logics and embeddings: Survey and
  perspective.
\newblock {\em arXiv preprint arXiv:2202.07412}, 2022.

\bibitem[\protect\citeauthoryear{Zhu \bgroup \em et al.\egroup
  }{2019}]{zhu-etal-2019-graph}
Hao Zhu, Yankai Lin, Zhiyuan Liu, Jie Fu, Tat-Seng Chua, and Maosong Sun.
\newblock Graph neural networks with generated parameters for relation
  extraction.
\newblock In {\em Proceedings of the 57th Annual Meeting of the Association for
  Computational Linguistics}, pages 1331--1339, Florence, Italy, July 2019.
  Association for Computational Linguistics.

\end{thebibliography}

\end{document}